\documentclass[10pt, a4paper]{article}
\usepackage{lrec2022} % this is the new LREC2022 Style
\usepackage{multibib}
\newcites{languageresource}{Language Resources}
\usepackage{graphicx}
\usepackage{stfloats}
\usepackage{tabularx}
\usepackage{array}
\usepackage{soul}
\usepackage[shortlabels]{enumitem}
\usepackage[T1]{fontenc}
\usepackage{titlesec}
\titleformat{\section}{\normalfont\large\bfseries\center}{\thesection.}{1em}{}
\titleformat{\subsection}{\normalfont\SmallTitleFont\bfseries\raggedright}{\thesubsection.}{1em}{}
\titleformat{\subsubsection}{\normalfont\normalsize\bfseries\raggedright}{\thesubsubsection.}{1em}{}
\renewcommand\thesection{\arabic{section}}
\renewcommand\thesubsection{\thesection.\arabic{subsection}}
\renewcommand\thesubsubsection{\thesubsection.\arabic{subsubsection}}
\usepackage{epstopdf}
\usepackage[utf8]{inputenc}

\usepackage{hyperref}
\usepackage{xstring}

\usepackage{color}
\usepackage{booktabs}

\makeatletter
\def\endthebibliography{%
  \def\@noitemerr{\@latex@warning{Empty `thebibliography' environment}}%
  \endlist
}
\makeatother

\newcommand{\ra}[1]{\renewcommand{\arraystretch}{#1}} % spacing between rows in booktabs table
\newcolumntype{P}[1]{>{\raggedright\arraybackslash\noindent}p{#1}} % noindent column type for booktabs

\title{Does Twitter know your political views? POLiTweets dataset and semi-automatic method for political leaning discovery}

\name{Joanna Baran, Michał Kajstura, Maciej Ziółkowski, Krzysztof Rajda} 

\address{Wroclaw University of Science and Technology \\
        Wyspiańskiego 27, 50-370 Wrocław, Poland \\
        \{joanna.baran, krzysztof.rajda\}@pwr.edu.pl, \{maciej.macron.ziolkowski, kajsturamichal\}@gmail.com\\}

\abstract{
Every day, the world is flooded by millions of messages and statements posted on Twitter or Facebook. Social media platforms try to protect users' personal data, but there still is a real risk of misuse, including elections manipulation. Did you know, that only 13 posts addressing important or controversial topics for society are enough to predict one's political affiliation with a 0.85 F1-score? To examine this phenomenon, we created a novel universal method of semi-automated political leaning discovery. It relies on a heuristical data annotation procedure, which was evaluated to achieve 0.95 agreement with human annotators (counted as an accuracy metric). We also present POLiTweets - the first publicly open Polish dataset for political affiliation discovery in a multi-party setup, consisting of over 147k tweets from almost 10k Polish-writing users annotated heuristically and almost 40k tweets from 166 users annotated manually as a test set. 
We used our data to study the aspects of domain shift in the context of topics and the type of content writers - ordinary citizens vs. professional politicians.
 \\ \newline \Keywords{political affiliation, political leaning, political profiling, Twitter dataset, political dataset} }

\begin{document}

\maketitleabstract

% basing on the user's likes distribution between politicians' tweets from different parties
%Additionally, we debate that textual data contained in tweets alone is enough for a deep learning model to classify a person's views with a satisfactory result that increases with the number of that user's collected posts.
%Such way of labeling data has proven to be a good indicator of the present-day citizen political profile and most importantly saves time and resources spent on manual annotation.

\section{ Introduction }
% Every day, the world is flooded by millions of messages and statements posted on social media platforms.
% The scope of application of such a large amount of data proved to be very broad thanks to development of machine learning techniques - from hate speech or fake profile detection, sentiment analysis, influencer marketing up to community detection~\cite{TK2021100395,SurveyMLSocialNetwork}. However, one of the most exciting, yet morally ambigous opportunities for scientists is to study the human behaviour on an unprecedented scale. Digital traces such as written texts, likes, retweets, followings, friends network etc. appear to be extremely useful in gaining knowledge about our personal lives~\cite{PrivateTraits}. This information is especially analyzed in a user's political leaning prediction task. Much research has been done in this area, mainly focusing on the two-party system present in the USA~\cite{6113114,Yan2017ThePO}. Binary political classification is less challenging than multi-party prediction problem which is still underestimated in published works, even though most democratic countries are characterized by a diverse political arena.

%Recently, social media has become one of the most popular platforms for expressing political opinions, allowing politicians, parties and researchers to discover voters' beliefs. 

Digital traces such as social media posts or interactions appear to be extremely powerful in gaining knowledge about our personal lives and predicting political beliefs \cite{PrivateTraits}. Automated analysis of electoral support on social media could effectively replace traditional surveys, being more cost-effective, allowing for examining a much larger portion of the population and giving more insights into voters' profiles. Unfortunately, as history has shown with the Cambridge Analytics example~\cite{CambridgeAnalytica}, it is easy to misuse personal data, even to manipulate election results. 
%This case revealed the issue of insufficient security of private data managed by platforms like Twitter and Facebook~.
Since then, social media companies have done a lot to strengthen the security of users' personal data \cite{CambridgeAnalitycaBlackBox}. Have they done enough? Is it still possible to easily access enough personal data to model users' political affiliation? 

To verify these questions, we propose a universal, semi-automatic political affiliation discovery method and a POLiTweets dataset - the novel political leaning discovery dataset, consisting of over 147k tweets from almost 10k Polish-writing users. As far as we know, this is the first published Polish dataset for predicting an individual's political orientation. 

The main objective of this article is to answer the following Research Questions: (\textbf{RQ1}) Is it possible to determine one's political affiliation using one's social media activity? (\textbf{RQ2}) How many social media posts are needed to accurately determine one's political leaning? (\textbf{RQ3}) Do professional politicians use the same language as their party supporters?

Our main contribution includes 3 points. Firstly, we designed a novel method of semi-automatic annotation of political leaning data. Secondly, we used so to collect the first Polish publicly open dataset for predicting political affiliation in a multi-party setup. Finally, we performed an analysis of the current political polarization in the Polish Twitter community.

% The remainder of this paper is organized as follows: Section 2 presents a literature review on the
% topic of predicting political affiliation; Section 3
% describes proposed method of data annotation; Section 4 describes details of POLiTweets dataset;
% Section 5 describes conducted experiments and summarizes the results; Section 6 discusses the results in terms of research
% questions; Section 7 presents conclusions from our works.

\section{ Related Work }
\label{sec:related_work}
Assessing political affiliation - which party a given user is a supporter - is the main challenge of constructing political affiliation detection dataset. There have been several works on labelling political orientation using Twitter or Facebook.

\textbf{Manual annotation.} Probably the most accurate method is to survey volunteers willing to declare their political opinion~\cite{PrivateTraits,PreotiucPietro2017BeyondBL} or manual annotation by a group of specialists who have access to publicly available user account information~\cite{ClassifyingPoliticalIsNotEasy,Samih2020AFT}. Unfortunately, both approaches are very time-consuming and practically unattainable for a large amount of data. For this reason, many automatic annotation methods have been proposed.

\textbf{Annotation by relations.} This approach assumes that the majority of the politicians we observe on Twitter reflect our views to some extent. The most popular method is to propagate party labels through analysis of followers and following users of politician accounts. \cite{IdeologicalExtremity,GOLBECK2014177,TwitterPsychologyUS,Barber2015BirdsOT}. However, the follower relationship can introduce some noise, given the existence of users observing opposing politicians (eg. to get a balanced opinion). \newcite{An2012VisualizingMB} provided annotation by exploiting the political bias of popular news media provider accounts in the USA and their mutual followers' group.

\textbf{Annotation by keywords.} The use of hashtags or keywords in tweets has been explored in some work as an indicator of user's affiliation, as exemplified by the Scottish Independence Referendum study of social media public opinions~\cite{TopicCentricClassification}. Similarly, \newcite{Tatman2017MAGAO} focused only on specific electoral slogans in users' bios or usernames.

\textbf{Annotations by interactions.}
These methods take advantage of users' interactions with the content they view. \newcite{IdentifyingUsersOpposingOpinions} introduced a semi-supervised retweet-based label propagation algorithm, based on the belief that retweeting a tweet may indicate endorsement of its content. 
%That was especially emphasized by \newcite{QuantifyingPoliticalLeaning} who formulated the prediction of a user's political leaning as a convex optimization problem given retweet and retweeter information.
Nevertheless, one can easily find tweets shared with a negative comment, retweeted to show disapproval. A more reliable premise is a user's "like" \cite{SocialMediaMicrotargeting}. Analysis of the likes distribution towards political parties' posts has proven to be very effective in this task and was verified by a manual survey in~\newcite{ParsimoniousData}. Like is a very strong signal of support towards the post creator and, in our opinion, the best choice to develop heuristics for assigning political labels to users' social media profiles. 

Finally, we need to point out that most of the published collections have focused on the two-party system, as is present in the USA~\cite{6113114,Yan2017ThePO} (binary classification setup). The multi-party prediction problem is far more challenging and yet still underestimated in research works, even though the majority of democratic countries are characterized by a diverse political arena. This highlights the importance of creating multi-party datasets for a more complex study of social media users.

\section{Political Afiliation Discovery Method }
\label{sec:methodology}
In this section, we describe the proposed semi-automatic method of political leaning discovery and test its quality. 

%First and main step of our method is to collect political afiliation dataset and annotate it in a heuristic way. Then the dataset is used in a standard way, to train a classifier, which can be applied to perform generalized analysis. 

\subsection{Method Steps}
\label{sec:method_steps}
Our approach to detect one's political affiliation goes as follows (Figure \ref{fig:pipeline}):

\begin{figure}[!h]
\begin{center}
\includegraphics[scale=0.8]{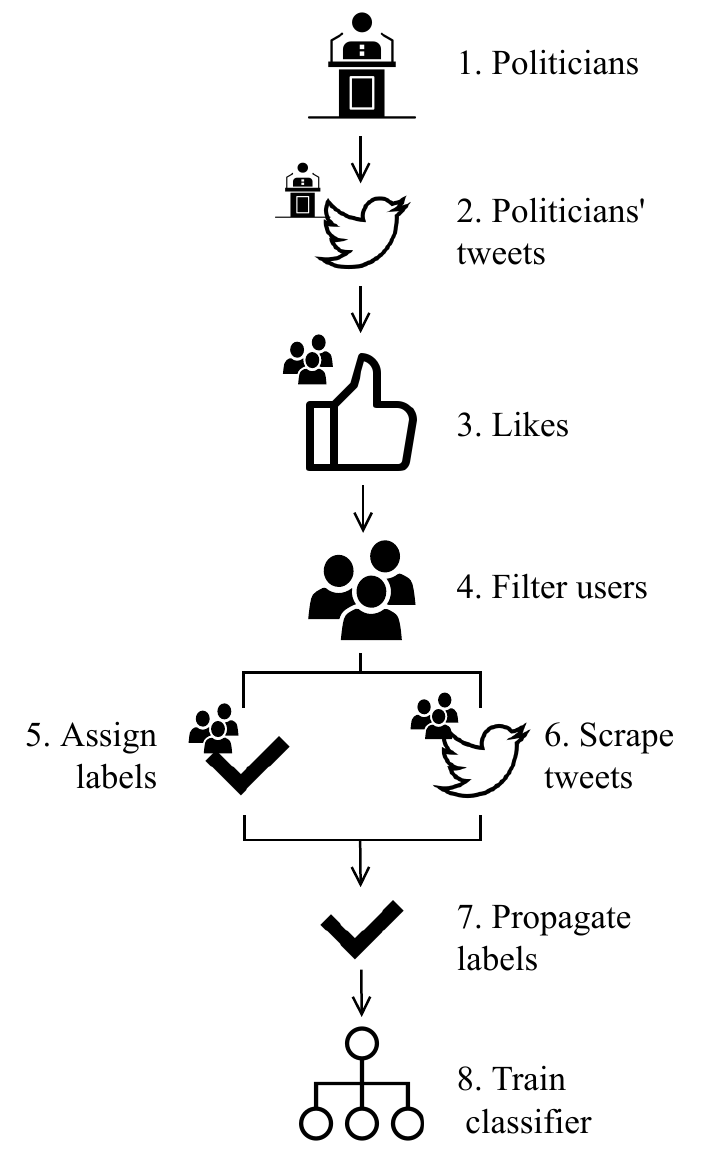} 
\caption{An illustration of the data acquisition pipeline}
\label{fig:pipeline}
\end{center}
\end{figure}

\begin{enumerate}[(1),wide, nosep]

\item Manually create the list of prominent political figures and their Twitter accounts in examined country/culture. Assign them a party label according to the politician's affiliation.

\item Aquire tweets posted from those accounts (we used an official Twitter API).

\item Save a list of social media users that liked collected politicians' tweets.

\item Filter out individuals with less than 10 likes to reduce the noise in acquired labels.

\item Count and group each users' likes by a political party. Choose the most frequent one as a label. Exclude inconclusive users with an equal number of likes for different parties for training purposes. For testing purposes, they could be manually annotated and included in the test set.
 
\item Collect selected users' tweets. Each scraped entry had to contain at least one predefined hashtag or keyword related to controversial topics - a post addressing this kind of politically loaded and divisive subject is more likely to reflect the political views of a person writing it~\newcite{OpposingViewsIncreasePoliticalPolarization}. 

\item Propagate obtained users' labels to all users' tweets, resulting in a heuristically annotated text classification dataset.

\item Train a text classifier, which can be then used for analysing the political affiliation of any provided text. It can be also applied to a user-level classification, aggregating individual posts' labels for each user, further increasing the accuracy of the method.
\end{enumerate}

\subsection{Heuristic Quality Evaluation}
\label{sec:ann_validation}
As the proposed method relies on a heuristic approach to data annotation, we ensured its performance by comparing it with human annotators. All users present in the test set were manually verified and labelled by three independent annotators using all available public data, such as posts, bio, followers network and uploaded images.
When a discrepancy occurred, a majority vote was taken to select a party name. Determining a person's political preferences based on social media posts is a highly subjective task, but annotators scored a high inter-annotator agreement of 0.74 Krippendorff's alpha coefficient. In total, we had 29 960 posts from 133 users for which we had human annotations.% and ones from the heuristic method.

Labels obtained using the proposed automatic labelling scheme matched the manual annotations with a \textbf{0.95 accuracy}, which confirms heuristics reliability.

\section{POLiTweets Dataset Summary} 
We used our proposed political leaning discovery method to collect a dataset consisting of \textbf{186 868 tweets} in Polish language, written by 9 837 Twitter users. Description and full hashtag list of controversial topics selected in method Step 6 are presented in Section~\ref{sec:hashtags_list}. Data were posted between March 2021 and January 2022. As labels set, we have chosen five main political groups with the greatest impact on the current Polish political scene - our selection was based on Parliament representation and election polls, more details in Section \ref{polls}.

\begin{figure}[!h]
\begin{center}
\includegraphics[scale=0.5]{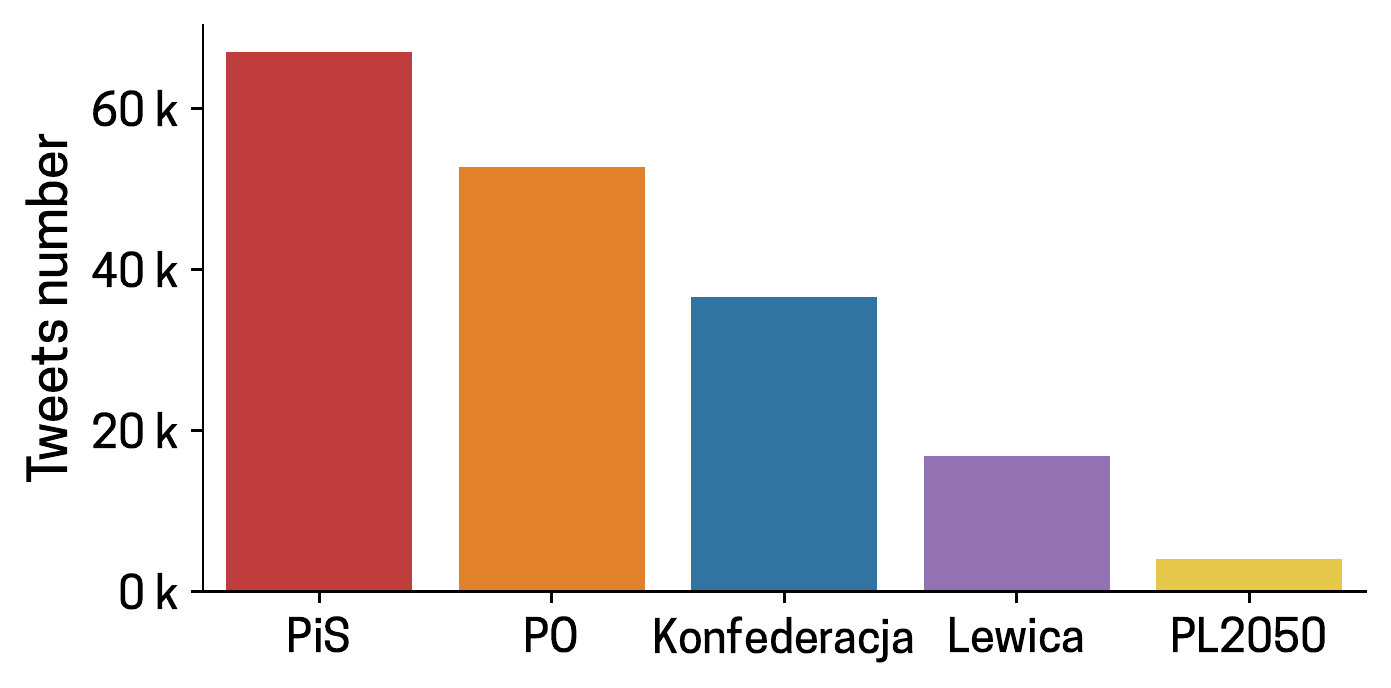} 
\caption{POLiTweets party label distribution}
\label{fig:party_dist}
\end{center}
\end{figure}

%The average number of tweets per user was 18, noting that for half of the community there were at most 6 posts per individual. 
The distribution of assigned labels was unbalanced (Figure~\ref{fig:party_dist}), which is quite consistent with the Polish political reality.
%We noted the high activity of \textit{Konfederacja} followers on the social media compared to their national support, probably caused by its supporters' demographic breakdown, with high participation of young people.
% - PiS and PO parties received the vast majority of votes in the last elections (PiS - 43.59\%, PO - 27.40\% in the 2019 general election). 

The previously manually annotated users and their tweets were taken as a test set (see Section~\ref{sec:ann_validation}). It contained only profiles with at least 15 posts per each. The rest of the data (over 147k tweets) was randomly split in a 9:1 ratio creating a training and validation split. We ensured that each entry has a minimum of 5 words that are not a hashtag, user mention or URL. The dataset is available in CSV files containing TweetID instead of explicit text, according to Twitter's guidelines about redistributing their content for scientific purposes. Test collection file appears twice due to different annotation sources - from heuristics method (\textit{heuristics-test}) and human annotators (\textit{manual-test}). We provide also a file with posts of the Polish parties' official Twitter accounts and their active politicians' list. Additionally, we prepared an \textit{ambiguous test set} of manually annotated tweets of 33 users for whom we skipped filtering by the ratio of likes in Step 5 of our method (Section~\ref{sec:method_steps}). We will refer to them as \textit{ambiguous} because discovering their political views is a much more difficult task even for a human annotator due to a more uniform distribution of those users' likes across political parties.

\section{Experiments and Results}
\label{sec:experiments}
The final step of our method was to train a classifier on the acquired tweets' textual data, without using any additional knowledge. We used it for analysis of current political polarization among the Polish Twitter community.

\textbf{Experimental setup.} As a classifier, we finetuned the base version of HerBERT - a BERT-based language model dedicated to Polish language \cite{HerBERT} - with a sequence classification head. The choice of this architecture was justified by its numerous effective applications in the NLP field. We considered it a good starting point for obtaining preliminary results on our dataset. The training took maximally 50 epochs with an early stopping patience parameter equal to 15 epochs. The batch size was set to 32 and the model was optimized using AdamW with a learning rate of 1e-5 along with a warmup linear scheduler. The preprocessing stage included the removal of hashtags, user mentions and URLs from posts to prevent data leakage. Due to the large imbalance of classes in the dataset, we applied a weighted sampler. 

\textbf{Model performace.} The text classifier achieved 0.64 micro F1-score on the \textit{manual-test} split. More detailed results are presented in Table~\ref{tab:f1_classes}.
\begin{table}[!h]
\centering
\begin{tabular}{@{}lccc@{}}
\toprule
Party name & Precision & Recall &  F1-score \\ \midrule
PiS  & 0.78 & 0.75      & 0.76 \\
PO  &  0.64 & 0.67    & 0.65 \\
Konfederacja & 0.35 & 0.36 & 0.36 \\
Lewica  & 0.18 & 0.26     & 0.21 \\
PL2050  & 0.31 & 0.07      & 0.12 \\ 
\midrule
Total & - & - & 0.64 \\
\bottomrule
\end{tabular}
\caption{Scores for particular classes}
\label{tab:f1_classes}
\end{table}

\begin{figure}[!h]
\begin{center}
\includegraphics[scale=0.38]{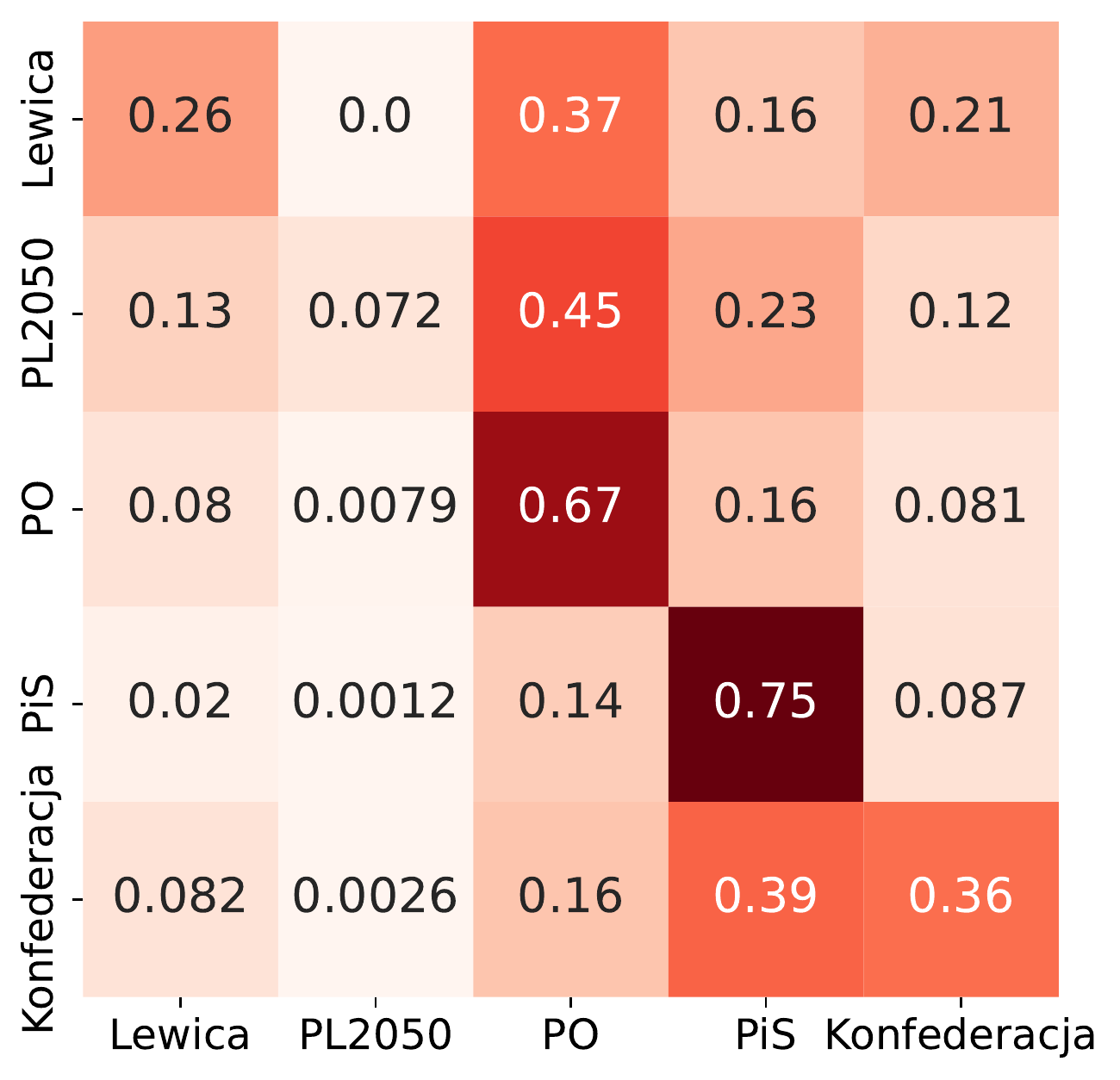} 
% \vspace*{-1\baselineskip}
\caption{F1-score confusion matrix}
\label{fig:confusion_matrix}
\end{center}
\end{figure}

\textbf{Classifier errors as Polish political scene descriptor.}
The confusion matrix, presented in Figure~\ref{fig:confusion_matrix},
reveals differences between political parties in Poland. Rows and columns are
ordered by the place in political spectrum, \textit{Lewica} being the furthest to the left and \textit{Konfederacja} on the rightmost side~\cite{publ286920}. The text classifier makes most errors between political groups with similar social and economic views. The lowest score is achieved for the centrist party - \textit{Polska2050}, which can be explained mostly by a strong class imbalance and hard to distinguish political opinions.

% \vspace*{-\baselineskip}

\textbf{Effects of Domain Shift.}
We investigated the effect of domain shift on our classifier by considering it from two perspectives. To examine writers' shift, the classifier trained previously on tweets from regular Twitter users (ordinary citizens) was used to predict party labels on professional politicians' posts, which were obtained in Step 2 of our method (Section  \ref{sec:methodology}). To study domain shift among topics, the training was carried out using the same experimental setup as stated above, but with a training set containing posts from 3 topics, leaving the last topic's posts as a test set.
\begin{table}[!h]
\centering
\begin{tabular}{lc}
\toprule
Domain-out  & F1-score \\ \midrule
Politicians          & 0.35             \\ 
Topic - abortion     & 0.40             \\ 
Topic - lexTVN       & 0.45             \\ 
Topic - EU \& CJEU   & 0.48             \\ 
Topic - The Polish Order & 0.46             \\ 
\bottomrule
\end{tabular}
\caption{Results for domain shift study. Each row states test set used}
\label{tab:domain_exp}
\end{table}

Results shown in the Table~\ref{tab:domain_exp} prove that the model's performance under domain shift is severely affected. The model tested on politicians' tweets achieved only 0.35 of micro averaged profile-level F1-score. A similar situation occurred for the experiment on topics - the highest result was 0.48 F1-score, compared to 0.65 without the domain shift scenario (Table \ref{tab:f1_classes}).

\section{Discussion}
\textbf{RQ1: Is it possible to determine one's political affiliation using one social media activity?} Yes, but we have to take into account the quality and balance of the input data. The research showed that the underrepresented parties were less recognizable by the classifier - model performance was the highest for more frequent political labels (like \textit{PiS} or \textit{PO} in the Polish case).

\begin{figure}[!h]
\begin{center}
\includegraphics[scale=0.4]{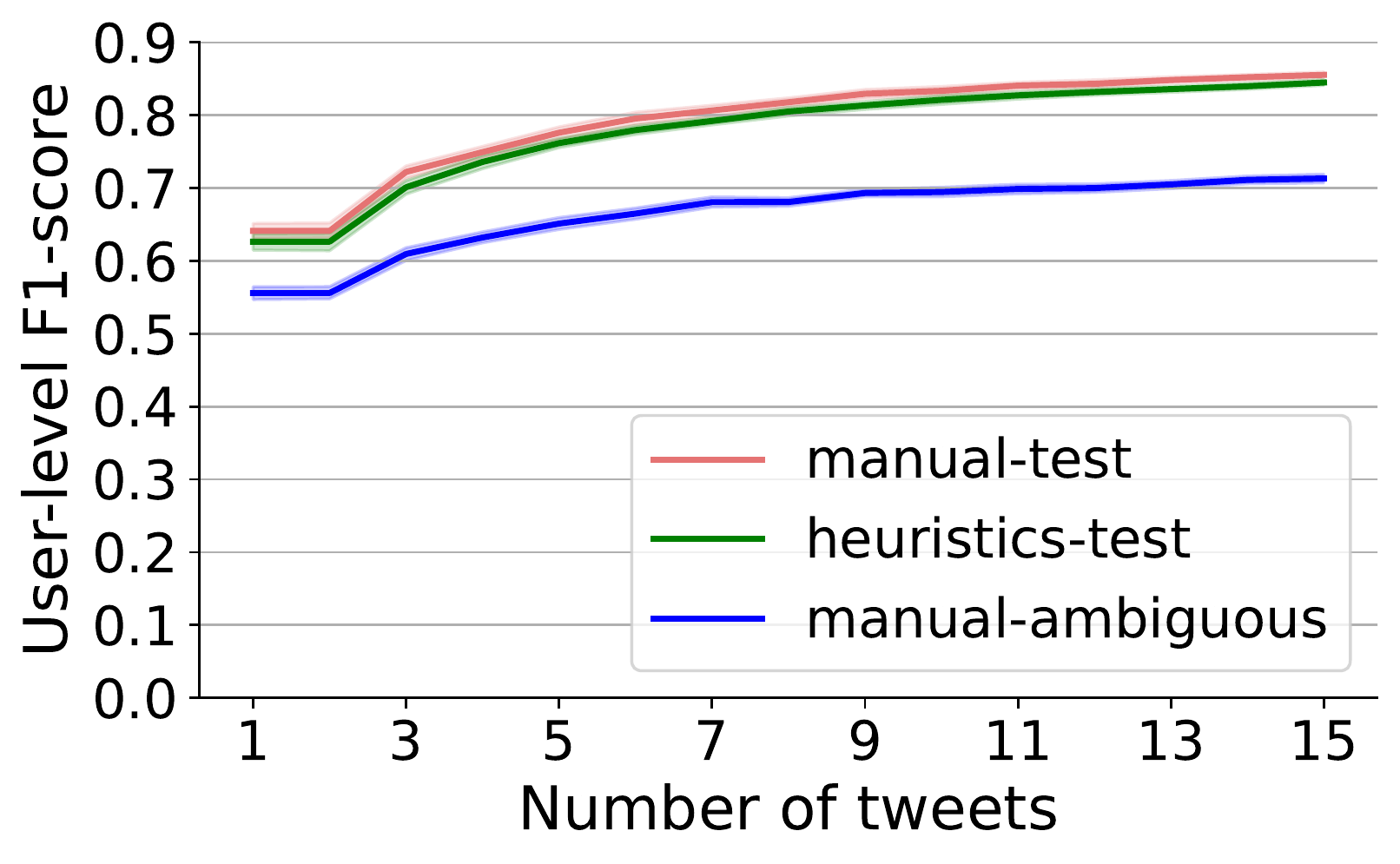}
\caption{Impact of the number of posts on model performance}
\label{fig:post_count}
\end{center}
\end{figure}

\textbf{RQ2: How many social media posts are needed to accurately determine one's political leaning?}
\label{sec:users_level}
To answer that question, we tested the classifier incrementally by increasing a subset of users' posts (from 1 to 15), choosing the most common label for the profile. We conducted evaluations on all 3 available test sets. The experiment was repeated 30 times with different tweet selection orders. Basing final predictions on more tweets drastically improves the performance of our method, which is presented in Figure~\ref{fig:post_count}. Accumulated predictions for well-defined users, for whom the number of collected likes was significantly higher for one party than the others, allowed to achieve a micro-averaged F1-score of 0.85 when used on 13 posts. For the more difficult user cases, the classifier scored significantly lower, but still above 0.5 F1-score.

\textbf{RQ3: Do professional politicians use the same language as their party supporters?}
Unfortunately, no. The classifier we chose proved to be sensitive to any changes in test data compared to training examples. It occurs not only in writer shift but also in the aspect of topic shift. This leads to the conclusion that despite the impressive results on in-domain texts, the deployment of such models should be carried out with special care. 

\textbf{Method advantages.}
The main advantage of our method is resource and time savings, compared to manual dataset annotation or traditional political affiliation surveying. Such surveys typically last for days to acquire a representative population, while our method can perform in only a few hours. It is also quite general - the data collection scheme can be applied to other social media platforms and any country or culture, due manual selection of controversial topics and seed profiles. Moreover, models trained on such data can also generalize to any kind of in-domain text, even those from users not actively liking any content produced by politicians.

\textbf{Method limitations.}
We assessed political leaning in a semi-automatic way. It requires choosing an initial set of politicians' accounts to scrape tweets from. It also needs a selection of some controversial, politically-loaded topics to filter out the discussion. Such topics depend mostly on country culture and current political debate and need to be chosen with care. 
%Our method also relies heavily on Twitter data, so accessing it can sometimes be challenging due to Twitter API limits.

\section{Conclusions and Future Works}
In this work, we proposed a semi-automatic, universal schema for political affiliation discovery, based on the heuristic method of data acquisition and annotation. Our approach proved to be highly effective, achieving a 0.95 user-level accuracy agreement when compared with the manually annotated dataset.
We also introduce POLiTweets - the first publicly available dataset for political leaning analysis in Polish, with almost 187k tweets from nearly 10k users. 
We proved that using tweets from popular and controversial topics, it is possible to associate Twitter users with political parties they support with sufficient confidence. Such knowledge may be exploited for microtargeting purposes, similar to how it was in the Cambridge Analytica scandal~\cite{CambridgeAnalytica}. Publishing even a few personal opinions on such topics may uncover users' political views - as our experiments showed, 13 posts are enough to classify political affiliation with a 0.85 F1-score.

Being aware of the high dependency on the data domain - topics and writers' type - we'd like to apply the domain adaptation techniques (eg. \newcite{ma-etal-2019-domain}) to support classifier stability. Also, more model architectures are needed to examine. We find it a good start for the follow-up work we tend to perform.

% \nocite{*}
\section{Bibliographical References}\label{reference}
%\label{main:ref}

\bibliographystyle{lrec2022-bib}
\bibliography{lrec2022-example}

% \section{Language Resource References}
% \label{lr:ref}
\bibliographystylelanguageresource{lrec2022-bib}
\bibliographylanguageresource{languageresource}

\section{Appendix}
\subsection{Used electoral polls}
\label{polls}
Our selection of political parties to label POLiTweets dataset was based on the results of the September 24-27, 2021 CAWI survey conducted on a nationwide and representative group of Polen~\cite{WNPSurveyCAWI}, which are as follows: Prawo i Sprawiedliwość (PiS, \textit{eng. Law and Justice}) - 38\%, Platforma Obywatelska (PO, \textit{eng. Civic Platform}) - 26\%, Polska 2050 (PL2050, \textit{eng. Poland 2050}) - 14\%, Konfederacja (\textit{eng. Confederation}) - 9\%, Lewica (\textit{eng. The Left}) - 8\%.

\subsection{Controversial topic selection} 
\label{sec:hashtags_list}
The period of data acquisition and conduction of our research was in 2021 and early 2022. At that time, the most divisive topics in Polish society with political overtones were:
\begin{enumerate}
    \item Abortion (\textit{pol. aborcja}) - due to the tightening of the abortion law and the resulting numerous strikes in the country,
    \item European Union, EU \& Court of Justice of the European Union, CJEU (\textit{pol. Unia Europejska, UE \& Trybunał Sprawiedliwości Unii Europejskiej, TSUE}) - in connection with the penalties imposed on Poland resulting from the extension of active mining in Turów and the expansion of anti-EU sentiments by the ruling party PiS,
    \item LexTVN - an amendment to the Broadcasting Act concerning the granting of broadcasting licenses to foreign entities,
    \item The Polish Order (\textit{pol. Polski Ład}) - the plan for the recovery of the Polish economy after the COVID-19 pandemic proposed from 2022, including new changes in tax law.
\end{enumerate}

\textbf{Topic distribution.} It is worth mentioning that in the POLiTweets dataset we obtained there were posts concerning several topics at the same time. We present detailed statistics in Table~\ref{tab:topic_keywords} along with a list of keywords we used to collect tweets via the API. Most of the data were acquired on \textit{EU \& CJEU} and \textit{LexTVN} - these were the main highlights of the 2021 year in Poland. The least amount of data comes from \textit{The Polish Order} topic, as the law came into effect in early 2022 when the collection of the dataset has ended.

\begin{table*}[ht]
\centering
\ra{1.3}
\begin{tabular}{@{}P{2.8cm}P{10.1cm}p{2.1cm}@{}}
\toprule
Topic & Keywords list & No. tweets \\ \midrule \
Abortion & aborcja, strajkkobiet, godek, czarnyprotest, AniJednejWięcej, prolife, BabiesLivesMatter, LegalnaAborcja, AborcjaJestOk   & 40 116 \\
EU \& CJEU & tsue, turów, polexit, konstytucja, zostajeMYwUE, MyZostajemy, NieWygasiciePolski, TrybunałKonstytucyjny, trybunał, UniaToMy, ZostajęwUnii, PolexitNow, ZostajemyWEuropie & 73 733 \\ 
LexTVN  & lextvn, tvn  & 63 609 \\ 
The Polish Order & PolskiŁad, Polski Ład, PolskiLad, Polski Lad & 5 277 \\
\bottomrule
\end{tabular}
\caption{Topics with descriptive keywords along with their number in the dataset}
\label{tab:topic_keywords}
\end{table*}

\end{document}